\documentclass{article}
\pdfoutput=1

\PassOptionsToPackage{numbers, compress}{natbib}

\usepackage[final]{neurips_2026}

\makeatletter
\def\@trackname{Main}
\let\@noticestring\@empty
\makeatother

\usepackage[utf8]{inputenc}
\usepackage[T1]{fontenc}
\usepackage{hyperref}
\usepackage{url}
\usepackage{booktabs}
\usepackage{amsfonts}
\usepackage{amsmath, amssymb}
\usepackage{nicefrac}
\usepackage{microtype}
\usepackage{xcolor}
\usepackage{graphicx}
\usepackage{caption}
\usepackage{enumitem}

\usepackage{tikz}
\usetikzlibrary{positioning, arrows.meta, calc}
\usepackage{pifont}
\usepackage{pgfplots}
\pgfplotsset{compat=1.18}
\usepgfplotslibrary{groupplots}

\definecolor{cKvpass}{RGB}{31,78,121}      % deep blue: KV-Fold
\definecolor{cFull}{RGB}{56,110,72}        % forest green: full attention / advantage
\definecolor{cStream}{RGB}{200,115,50}     % rust orange: StreamingLLM
\definecolor{cIso}{RGB}{135,135,140}       % gray: isolated chunks
\definecolor{cNeedle}{RGB}{56,142,90}      % bright green: needle / success
\definecolor{cCache}{RGB}{215,160,75}      % warm amber: KV cache material
\definecolor{cAccent}{RGB}{60,60,68}       % near-black slate: text & axes
\definecolor{cMuted}{RGB}{225,225,228}     % light gray: gridlines & structure

\author{%
  Alireza Nadali \thanks{Equal Contribution} \\
  Department of Computer Science\\
  University of Colorado Boulder\\
  Boulder, Colorado, 80301 \\
  \texttt{a\_nadali@colorado.edu} \\
   \And
  Patrick Cooper\footnotemark[1] \\
  Department of Computer Science\\
  University of Colorado Boulder\\
  Boulder, Colorado, 80301 \\
   \texttt{patrick.cooper@colorado.edu} \\
  \AND
 Ashutosh Trivedi \\
  Department of Computer Science\\
  University of Colorado Boulder\\
  Boulder, Colorado, 80301 \\
   \texttt{ashutosh.trivedi@colorado.edu} \\
   \And
  Alvaro Velasquez \\
  Department of Computer Science\\
  University of Colorado Boulder\\
  Boulder, Colorado, 80301 \\
   \texttt{alvaro.velasquez@colorado.edu} \\
}

\title{KV-Fold: One-Step KV-Cache Recurrence for Long-Context Inference}

\begin{document}
\maketitle

\begin{abstract}
We introduce \emph{KV-Fold}, a simple, training-free long-context
inference protocol that treats the key--value (KV) cache as the
accumulator in a left fold over sequence chunks. At each step, the
model processes the next chunk conditioned on the accumulated cache,
appends the newly produced keys and values, and passes the enlarged
cache forward; the same one-step update is applied repeatedly,
analogous to \texttt{foldl} in functional programming. Building on
the KV cache concatenation primitive introduced for latent
multi-agent communication, we repurpose it as a chunk-to-chunk
recurrence for long-context inference. When processing chunk $t$,
the model attends to the KV cache carried from earlier chunks as a
prefix, reusing its internal state across segments without modifying
or retraining the model.
% We introduce \emph{KV-Fold}, a simple, training-free
% mechanism that turns a pretrained transformer into a recurrent model
% and extends its effective context beyond a single forward pass. When
% processing chunk $t$, the model attends to the full key--value (KV)
% cache from chunk $t{-}1$ as a prefix, reusing its internal state
% across segments.
Despite its simplicity, the induced recurrence is stable: per-step
drift rises briefly and then saturates into a flat plateau that
persists across deep chains. This plateau is insensitive to a
$10{,}000\times$ change in numerical precision, robust across chunk
sizes, and consistent across model families. At the task level,
KV-Fold preserves exact information over long distances.
On a needle-in-a-haystack benchmark, it achieves $100\%$ exact-match
retrieval across $152$ trials spanning contexts from $16$K to $128$K
tokens and chain depths up to $511$ on Llama-3.1-8B, while remaining
within the memory limits of a single 40GB GPU.
Compared to streaming methods, which trade fidelity for bounded
memory, KV-Fold maintains long-range retrieval while
operating as a sequence of tractable forward passes. Overall, our
results show that frozen pretrained transformers already support a
stable form of KV-cache recurrence, providing a practical route to
long-context inference without architectural changes or training.
\end{abstract}

\section{Introduction}
Large pretrained transformers~\cite{vaswani2017attention} have become the default architecture for language modeling, code generation, and long-form reasoning. 
However, their ability to use context remains constrained by the \emph{attention budget} of a single forward pass. 
A token can attend only to the tokens represented within the current attention computation, while the memory cost of full attention grows quadratically with sequence length. As a result, applications that naturally generate long streams of information, such as repository-scale code assistance, longitudinal medical records, customer preference modeling over extended interaction histories, and monitoring of sensitive defense operations, quickly exceed the practical limits of ordinary inference. 
This raises a natural question: \emph{can a pretrained transformer carry its internal state across successive chunks, turning long-context inference into a recurrence rather than a single monolithic forward pass?}

Although recent models have pushed context windows into the
hundreds of thousands or even millions of tokens, the restriction
remains binding in practice. Even at this scale, real workloads can
exceed the available context: large codebases such as DOE systems
like E3SM, FLASH, and NWChem span millions of lines of code, with
each line corresponding to roughly ten or more tokens. Processing
such sequences still requires truncation, repeated re-encoding, or
approximations that alter the attention computation. A large body
of work has sought to overcome this limitation. Streaming methods maintain bounded inference memory
by retaining a small set of attention sinks together with a sliding
window of recent tokens~\cite{streamingllm,han2024lminfinite}.
KV-cache compression methods attempt to preserve useful past
information while evicting, pruning, or quantizing parts of the
cache~\cite{zhang2023h2o,liu2023scissorhands,li2024snapkv,cai2024pyramidkv,ge2024fastgen,liu2024kivi,hooper2024kvquant}.
Other approaches introduce recurrent memory or learned compression
mechanisms~\cite{dai2019transformerxl,rae2019compressive,wu2022memorizing,bulatov2022rmt,hutchins2022blockrecurrent,munkhdalai2024infini},
modify positional encodings or attention patterns, or replace attention
with state-space dynamics~\cite{gu2022s4,gu2023mamba}. These methods
have produced impressive progress, but they also expose a persistent
trade-off: they either modify the model, require training or
fine-tuning, compress the past into a lossy state, approximate the
attention computation, or discard old tokens entirely. In settings
where exact recall matters, such as recovering an identifier from a
long log or preserving a factual detail introduced much earlier in a
document, these trade-offs can be unacceptable.

This paper explores a simpler possibility: the key--value (KV) cache
of a pretrained transformer can itself serve as a recurrent state. 
In a decoder-only transformer, the KV cache stores layer-wise
representations of previous tokens, which later tokens access through
attention. Although this cache is usually treated as a serving
optimization, it is also a structured record of the model's past
computation. 
Recent work on latent multi-agent communication~\cite{zou2025latentmas}
showed that one transformer pass can attend to another pass's KV cache
as a prefix, allowing information transfer directly through latent
state. We repurpose this primitive for long-context inference within a
single pretrained model.

We introduce \emph{KV-Fold}. A long sequence is divided into
chunks. When processing chunk $t$, the model attends to the accumulated
KV cache from earlier chunks as a prefix, produces new keys and values
for the current chunk, and passes the enlarged cache to chunk
$t{+}1$. This turns standard transformer inference into a simple
chunk-to-chunk recurrence in which the KV cache plays the role of the
accumulator in a left fold: a single one-step update is applied
repeatedly over the chunk sequence, analogous to \texttt{foldl} in
functional programming. The model parameters remain unchanged, no
special memory tokens are introduced, and no fine-tuning is required.
The shift is purely operational: instead of a single monolithic
full-context forward pass, the model processes the sequence as a chain
of manageable forward passes, carrying its state forward through the
KV cache. This comes with a tradeoff: the cache grows linearly with
sequence length, increasing memory use and per-step latency relative
to streaming approaches.

The main empirical finding is that this recurrence is stable. If each
chunk boundary introduced independent error, one would expect the
difference from full attention to grow with chain depth. Instead, we
observe a short transient followed by a flat plateau: per-step drift
rises in the first few transitions and then stabilizes, remaining
essentially unchanged even across deep chains. This plateau is
insensitive to a $10{,}000\times$ change in numerical precision,
robust across chunk sizes, and consistent across transformer
families. In other words, KV-Fold does not behave like
accumulated numerical error; it moves the model into a slightly
shifted but stable attention regime.

\begin{figure}[t]
\caption{\textbf{Headline result.} \emph{Left:} on Llama-3.1-8B at $T=128$K, KV-Fold holds $100\%$ exact-match retrieval at every tested distance through chain depth $511$, while StreamingLLM falls to $0\%$ once the needle exits its $1024$-token cache. \emph{Right:} peak GPU memory grows linearly at $0.13$~KB per token; the chain at $T=128$K, depth $511$ completes in $171$~s.}
\label{fig:hero}
\centering
\begin{tikzpicture}
\begin{groupplot}[
  group style={group size=2 by 1, horizontal sep=1.7cm},
  width=0.46\linewidth, height=4.9cm,
  axis line style={draw=cAccent, line width=0.5pt},
  tick style={draw=cAccent, line width=0.4pt},
  major grid style={line width=0.2pt, draw=cMuted},
  label style={font=\small, color=cAccent},
  tick label style={font=\footnotesize, color=cAccent},
  title style={font=\small\bfseries, color=cAccent, yshift=2pt, align=center},
  every axis plot/.append style={line width=1.4pt},
]

% Panel A: Retrieval cliff
\nextgroupplot[
  title={Retrieval at $T = 128$K},
  xlabel={Needle-to-question distance (chain transitions)},
  ylabel={Exact-match retrieval},
  xmode=log, log basis x=10,
  xmin=0.7, xmax=900,
  ymin=-7, ymax=115,
  xtick={1,31,255,511},
  xticklabels={$1$, $31$, $255$, $511$},
  ytick={0,25,50,75,100},
  yticklabel={\pgfmathprintnumber{\tick}\%},
  ymajorgrids,
  legend style={
    at={(0.97,0.50)}, anchor=east,
    font=\footnotesize, draw=cMuted,
    fill=white, fill opacity=0.96, text opacity=1,
    inner sep=3pt, row sep=1pt,
  },
  legend cell align=left,
]
\addplot[color=cKvpass, mark=*, mark size=2.8pt, mark options={fill=cKvpass, draw=white, line width=0.7pt}]
  coordinates {(1, 100) (31, 100) (255, 100) (511, 100)};
\addlegendentry{KV-Fold}
\addplot[color=cStream, mark=square*, mark size=2.5pt, mark options={fill=cStream, draw=white, line width=0.6pt}]
  coordinates {(1, 100) (31, 0) (255, 0) (511, 0)};
\addlegendentry{StreamingLLM}

% Panel B: Memory scaling
\nextgroupplot[
  title={Scaling on a $40$\,GB A100},
  xlabel={Total context $T$ (tokens)},
  ylabel={Peak GPU memory (GB)},
  xmin=22, xmax=140,
  ymin=10, ymax=46,
  xtick={32,64,96,128},
  xticklabels={$32$K, $64$K, $96$K, $128$K},
  ytick={10,20,30,40},
  ymajorgrids,
]
% A100 ceiling
\addplot[color=cAccent, dashed, line width=0.7pt, forget plot]
  coordinates {(22,40) (140,40)};
\node[anchor=south east, font=\scriptsize, color=cAccent]
  at (axis cs:138, 40.3) {A100 $40$\,GB};

% Memory data
\addplot[color=cKvpass, mark=*, mark size=2.8pt, mark options={fill=cKvpass, draw=white, line width=0.7pt}]
  coordinates {(32, 21.0) (64, 25.86) (96, 30.72) (128, 35.57)};

\node[anchor=north west, align=left, font=\scriptsize, color=cKvpass, fill=white, fill opacity=0.92, text opacity=1, inner sep=1.5pt]
  at (axis cs:35, 20) {$100\%$ retrieval at\\all $T$ ($152/152$ trials)};

\end{groupplot}
\end{tikzpicture}
\end{figure}
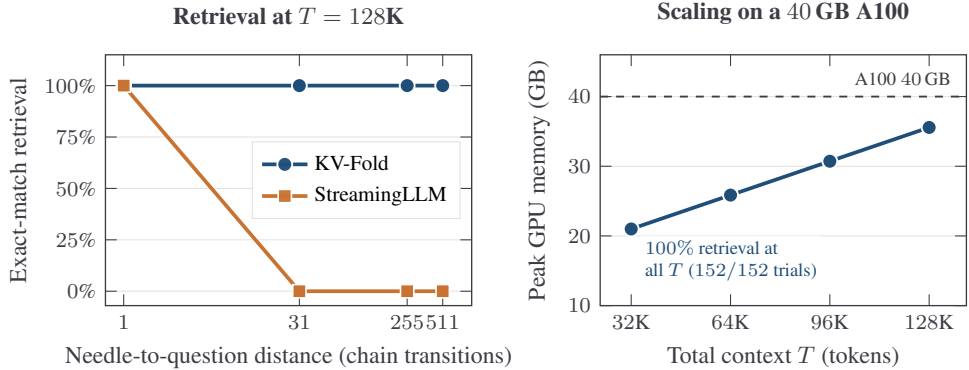

Notably, this stable regime preserves useful long-range
information. Because the KV state is not compressed, earlier
tokens remain available for content-based addressing. On a
needle-in-a-haystack benchmark spanning $T {\in} \{16\text{K},
32\text{K}, 64\text{K}, 96\text{K}, 128\text{K}\}$ tokens, KV-Fold
recovers $100\%$ of inserted facts at every tested
distance, including chain depth $511$ on Llama-3.1-8B at the full
$128$K native context window.

The method occupies a distinct point on the memory-versus-retrieval
trade-off. Compared to full attention, peak working memory is
dramatically lower: at $T = 128$K on a $40$~GB A100, a single
full-attention forward is infeasible (its attention-scores matrix
alone is ${\sim}1$~TB), while KV-Fold fits within the
hardware budget at $35.6$~GB peak and completes in $171$~s.
Compared to streaming methods, peak memory is substantially higher:
StreamingLLM~\cite{streamingllm} holds a constant $1024$-token
cache ($16.6$~GB peak at $T {=} 128$K) while KV-Fold's
cache grows linearly to $17.18$~GB for similar setting. The
extra memory pays for retrieval: KV-Fold maintains
$100\%$ at every tested distance, while StreamingLLM loses access
once the needle exits its $1024$-token window
(Figure~\ref{fig:hero}).
% Figure~\ref{fig:hero} shows the contrast with
% StreamingLLM~\cite{streamingllm} directly: KV-Fold
% maintains $100\%$ retrieval at every tested distance, while
% StreamingLLM falls to $0\%$ once the needle exits its $1024$-token
% cache.

This paper introduces \emph{KV-Fold}, a training-free
protocol for long-context inference with pretrained transformers, and
provides a systematic empirical characterization of its behavior.
Our contributions are as follows:
\begin{itemize}[topsep=2pt, itemsep=3pt, parsep=0pt, leftmargin=1.5em]
  \item We propose KV-Fold, a simple inference-time protocol
  that turns a frozen pretrained transformer into a recurrent
  long-context model by carrying the accumulated KV cache across
  chunks as the accumulator in a left fold, without architectural
  changes, special memory tokens, or fine-tuning.

  \item We show that the induced recurrence is stable: its drift
  relative to full attention rises briefly and then saturates into a
  flat plateau, rather than accumulating with chain depth.

  \item We show that this plateau is structural rather than numerical:
  it is largely unchanged under a $10{,}000\times$ increase in numerical
  precision, consistent across model families, and robust across a wide
  range of chunk sizes.

  \item We demonstrate that KV-Fold preserves exact
  long-range information, matching full-attention retrieval on
  needle-in-a-haystack benchmarks through contexts up to $128$K tokens
  and chain depths up to $511$.

  \item We characterize its practical trade-offs relative to streaming
  methods such as StreamingLLM: streaming achieves bounded-memory and
  faster runtime, but loses retrieval once content exits the window,
  whereas KV-Fold preserves retrieval by retaining the full
  KV history with linear memory.
\end{itemize}

\noindent\textbf{Overview of questions.}
Taken together, these results raise a set of concrete questions about
the behavior of the recurrence. Does error accumulate with chain
depth, or does it settle into a bounded regime? If drift appears, is it
driven by numerical precision or by a structural shift in the attention
computation? How robust is this behavior across architectures and
operational choices such as chunk size? Does the recurrence preserve
task-level information, in particular the ability to retrieve exact
facts introduced far back in the sequence? Finally, how does the method
scale in practice, in terms of memory, compute, and achievable context
length on real hardware? We answer each of these questions empirically
in the sections that follow.

\noindent\textbf{Organization.}
Section~\ref{sec:method} introduces KV-Fold.
Section~\ref{sec:depth} studies the stability of the recurrence and
characterizes drift as a function of chain depth. 
Section~\ref{sec:precision} examines whether this drift is numerical or
structural, and robustness across architectures and chunk sizes.
Section~\ref{sec:needle} measures task-level retrieval, and
Section~\ref{sec:scale} analyzes scaling behavior in memory, compute,
and context length.

\section{KV-Fold}
\label{sec:method}

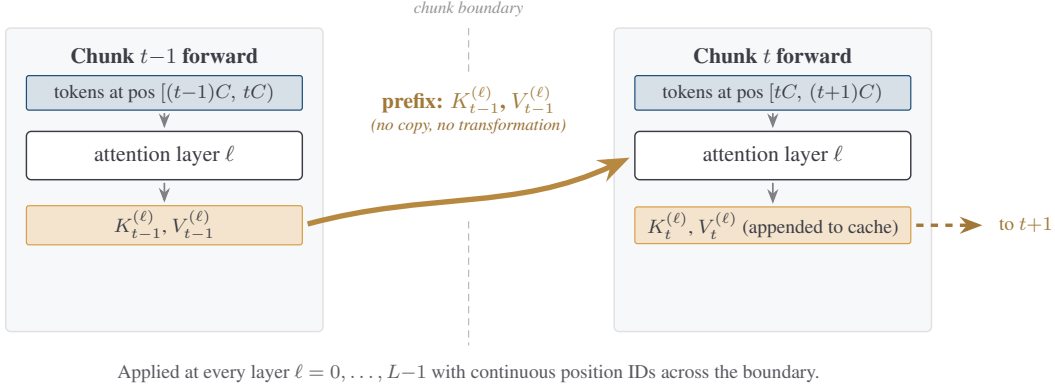
\begin{figure}[t]
\centering
\resizebox{\linewidth}{!}{%
\begin{tikzpicture}[
  font=\small,
  >=Stealth,
  chunkbox/.style={
    rectangle, rounded corners=2pt,
    draw=cMuted, line width=0.6pt,
    fill=cKvpass!4,
    minimum width=4.6cm, minimum height=4.4cm,
    inner sep=4pt
  },
  tokens/.style={
    rectangle, rounded corners=1pt,
    draw=cKvpass, line width=0.5pt,
    fill=cKvpass!18,
    minimum width=4.0cm, minimum height=0.5cm,
    inner sep=2pt, font=\footnotesize, text=cAccent
  },
  attn/.style={
    rectangle, rounded corners=2pt,
    draw=cAccent, line width=0.6pt,
    fill=white,
    minimum width=4.0cm, minimum height=0.7cm,
    inner sep=2pt, font=\small, text=cAccent
  },
  kvbox/.style={
    rectangle, rounded corners=1pt,
    draw=cCache, line width=0.5pt,
    fill=cCache!28,
    minimum width=4.0cm, minimum height=0.5cm,
    inner sep=2pt, font=\footnotesize, text=cAccent
  },
  iarr/.style={->, draw=cAccent!65, line width=0.7pt, shorten >=1pt, shorten <=1pt},
  carry/.style={->, line width=2.2pt, draw=cCache!85!black,
    shorten >=2pt, shorten <=2pt},
  outflow/.style={->, line width=1.4pt, draw=cCache!75!black, dashed,
    shorten >=2pt, shorten <=2pt},
  boundary/.style={densely dashed, draw=cAccent!35, line width=0.5pt},
]
\node[chunkbox] (left) at (0, 0) {};
\node[anchor=north, font=\small\bfseries, color=cAccent]
  at ($(left.north) + (0, -0.18)$) {Chunk $t{-}1$ forward};
\node[tokens] (Lin) at ($(left.north) + (0, -0.95)$)
  {tokens at pos $[(t{-}1)C,\, tC)$};
\node[attn] (Lattn) at ($(Lin.south) + (0, -0.65)$)
  {attention layer $\ell$};
\node[kvbox] (Lkv) at ($(Lattn.south) + (0, -0.65)$)
  {$K^{(\ell)}_{t-1}$, $V^{(\ell)}_{t-1}$};
\draw[iarr] (Lin) -- (Lattn.north);
\draw[iarr] (Lattn.south) -- (Lkv.north);

\node[chunkbox, right=4.2cm of left] (right) {};
\node[anchor=north, font=\small\bfseries, color=cAccent]
  at ($(right.north) + (0, -0.18)$) {Chunk $t$ forward};
\node[tokens] (Rin) at ($(right.north) + (0, -0.95)$)
  {tokens at pos $[tC,\, (t{+}1)C)$};
\node[attn] (Rattn) at ($(Rin.south) + (0, -0.65)$)
  {attention layer $\ell$};
\node[kvbox] (Rkv) at ($(Rattn.south) + (0, -0.65)$)
  {$K^{(\ell)}_t$, $V^{(\ell)}_t$ (appended to cache)};
\draw[iarr] (Rin) -- (Rattn.north);
\draw[iarr] (Rattn.south) -- (Rkv.north);

\path let \p1 = ($(left.east)!0.5!(right.west)$) in
  coordinate (mid) at (\p1);
\draw[boundary] ($(mid) + (0, 2.2)$) -- ($(mid) + (0, 1.5)$);
\draw[boundary] ($(mid) + (0, -0.6)$) -- ($(mid) + (0, -2.4)$);
\node[anchor=south, font=\scriptsize\itshape, color=cAccent!55]
  at ($(mid) + (0, 2.25)$) {chunk boundary};

\draw[carry] (Lkv.east) to[out=15, in=205] (Rattn.west);
\node[font=\small\bfseries, color=cCache!75!black]
  at ($(mid) + (0, 1.15)$) {prefix: $K^{(\ell)}_{t-1}$, $V^{(\ell)}_{t-1}$};
\node[font=\scriptsize\itshape, color=cCache!70!black]
  at ($(mid) + (0, 0.78)$) {(no copy, no transformation)};

\draw[outflow] (Rkv.east) to[out=0, in=180] ($(Rkv.east) + (1.1, 0)$);
\node[anchor=west, font=\footnotesize, color=cCache!75!black]
  at ($(Rkv.east) + (1.15, 0.0)$) {to $t{+}1$};

\node[anchor=north, font=\footnotesize, color=cAccent!85, align=center]
  at ($(left.south)!0.5!(right.south) + (0, -0.35)$)
  {Applied at every layer $\ell = 0, \ldots, L{-}1$ with continuous
   position IDs across the boundary.};
\end{tikzpicture}%
}
\caption{\textbf{KV-Fold.} Queries in chunk $t$ attend to the KV
cache from earlier chunks (amber) as a prefix; chunk $t$'s KV states
are then appended and carried forward. No compression or additional
parameters are introduced.}
\label{fig:mechanism}
\end{figure}

We adopt the chunked-attention setup standard in the long-context
literature.
A long sequence of length $T$ is divided into chunks of length $C$,
yielding $N = T/C$ chunks indexed $0, 1, \ldots, N{-}1$.
Each chunk is processed as a single forward pass through the model.
%\noindent \textbf{KV-Fold.}
At chunk $t$, we run a forward pass on the input chunk
$\mathbf{x}_t = [x_{tC}, x_{tC+1}, \ldots, x_{(t+1)C-1}]$
(boldface $\mathbf{x}_t$ denotes the length-$C$ chunk vector;
plain $x_i$ denotes the individual token at absolute position $i$),
and use the previous chunk's KV cache as the prefix.
Concretely, for each transformer layer $\ell$, the attention
computation at chunk $t$ produces queries $Q^{(\ell)}_t$ from the new
tokens and uses keys
$K^{(\ell)}_{0:t} = [K^{(\ell)}_0; K^{(\ell)}_1; \ldots; K^{(\ell)}_{t-1}; K^{(\ell)}_t]$
and analogously concatenated values
$V^{(\ell)}_{0:t}$, where $K^{(\ell)}_s$ and $V^{(\ell)}_s$ are the
keys and values produced when chunk $s$ was processed.
Position identifiers for the new tokens are continued from the
absolute sequence position $tC$, so the positional embedding RoPE~\cite{su2024roformer}
rotations are aligned with the original sequence.
This is the LatentMAS Equation~4 primitive applied across consecutive
chunks instead of across distinct agents.
Figure~\ref{fig:mechanism} illustrates one chunk-to-chunk transition:
chunk $t{-}1$'s K and V are passed unchanged into chunk $t$'s attention
as prefix, with continuous position IDs across the boundary.

\noindent \textbf{A recurrent view.}
This procedure can be interpreted as recurrence over chunks. Let
$(\mathcal{K}^{(t)}, \mathcal{V}^{(t)})$ denote the accumulated KV
cache after chunk $t$. Then
\[
(\mathcal{K}^{(t)}, \mathcal{V}^{(t)}) =
\mathcal{F}_\theta\big((\mathcal{K}^{(t-1)}, \mathcal{V}^{(t-1)}),
\mathbf{x}_t\big),
\]
where $\mathcal{F}_\theta$ is the standard transformer forward pass.
Equivalently, the full inference can be written as a left fold over
the chunk sequence,
\[
(\mathcal{K}^{(N-1)}, \mathcal{V}^{(N-1)}) =
\texttt{foldl}\big(\mathcal{F}_\theta,\;
(\emptyset, \emptyset),\;
[\mathbf{x}_0, \mathbf{x}_1, \ldots, \mathbf{x}_{N-1}]\big),
\]
with the KV cache playing the role of the accumulator and the same
one-step update applied at every step.
Unlike classical recurrent models, this state is not compressed into
a fixed-dimensional vector; it grows with the sequence and retains
structured representations of past tokens, allowing continued
content-based addressing through attention.

\noindent \textbf{Three reference conditions.}
For every experiment we measure three quantities per chunk:
\begin{itemize}[noitemsep, topsep=0.2pt, leftmargin=0pt]
\setlength{\itemsep}{2pt}
  \item \textsc{full}: the next-token negative log-likelihood under
    a single full-attention forward pass over the entire $T$-token
    sequence. This is the ceiling.
  \item \textsc{isolated}: Negative Log Likelihood (NLL) when each chunk is processed in
    isolation, with no prefix. This is the floor: each chunk sees
    no context from earlier chunks.
  \item \textsc{kv-fold}: NLL when each chunk is processed with the
    accumulated KV-cache prefix described above.
\end{itemize}

We define the per-depth drift as
$\textrm{drift}(d) = \textsc{NLL}_{\textsc{kv-fold}}(d) - \textsc{NLL}_{\textsc{full}}(d)$
and the per-depth recurrence advantage as
$\textrm{advantage}(d) = \textsc{NLL}_{\textsc{isolated}}(d) - \textsc{NLL}_{\textsc{kv-fold}}(d)$,
where $d$ is the chain depth (chunk index minus 1).
We use the validation split of PG-19~\cite{rae2019compressive}, a
benchmark of long-form prose, sampled at random with a fixed seed.
Each window starts with 200 tokens into a document and spans the next $T$
tokens.
All NLL values are computed in bfloat16 unless stated otherwise.

\section{Stability of the recurrence}
\label{sec:depth}

If KV-Fold simply accumulated error at each step, drift
would grow steadily with depth. Instead, we observe a short transient
followed by a stable plateau. On Qwen2.5-7B-Instruct~\cite{qwen25}
with $T = 16{,}384$ and $C = 256$ (depths $0$--$63$), drift rises
during the first few transitions, saturates by depth seven, and
remains essentially unchanged thereafter. Between depths $15$ and
$60$, the total change is just $-0.0003$~nats, well within
window-to-window noise ($N=42$). The central empirical result is
therefore not growing error, but bounded drift: after a small initial
shift at early chunk boundaries, further chaining does not move the
system farther from the full-attention regime.

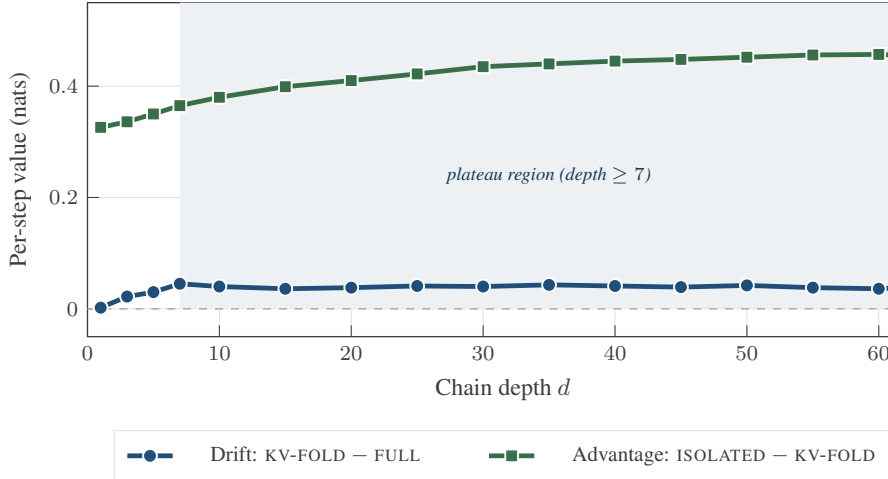
\begin{figure}[t]
\centering
\begin{tikzpicture}
\begin{axis}[
  width=0.9\textwidth,
  height=6.0cm,
  xlabel={Chain depth $d$},
  ylabel={Per-step value (nats)},
  xmin=0, xmax=63,
  ymin=-0.05, ymax=0.55,
  ymajorgrids,
  xmajorgrids,
  major grid style={line width=0.2pt, draw=cMuted},
  axis line style={draw=cAccent, line width=0.5pt},
  tick style={draw=cAccent, line width=0.4pt},
  label style={font=\small, color=cAccent},
  tick label style={font=\footnotesize, color=cAccent},
  legend style={
    font=\footnotesize, color=cAccent,
    draw=cMuted, fill=white, fill opacity=0.96, text opacity=1,
    at={(0.5,-0.28)}, anchor=north,
    legend columns=2,
    /tikz/every even column/.append style={column sep=0.8cm},
    inner sep=4pt, column sep=0.4cm,
  },
  legend cell align=left,
  every axis plot/.append style={line width=1.8pt},
]
  % Plateau region shading (background layer, drawn first)
  \addplot[fill=cKvpass!8, draw=none, forget plot]
    coordinates {(7,-0.05) (7,0.55) (63,0.55) (63,-0.05)} \closedcycle;
  \node[anchor=center, font=\scriptsize\itshape, color=cKvpass!70!black]
    at (axis cs:35, 0.24) {plateau region (depth $\geq 7$)};

  % Zero reference line
  \addplot[mark=none, dashed, color=cAccent!40, line width=0.6pt, forget plot]
    coordinates {(0, 0) (63, 0)};

  % Drift curve (in-front, thicker line, larger markers with white outlines)
  \addplot[color=cKvpass, mark=*, mark size=2.6pt,
           mark options={fill=cKvpass, draw=white, line width=0.7pt}]
    coordinates {
      (1, 0.002) (3, 0.022) (5, 0.030) (7, 0.045) (10, 0.040)
      (15, 0.036) (20, 0.038) (25, 0.041) (30, 0.040) (35, 0.043)
      (40, 0.041) (45, 0.039) (50, 0.042) (55, 0.038) (60, 0.036)
      (63, 0.040)
    };
  \addlegendentry{Drift: \textsc{kv-fold} $-$ \textsc{full}}

  % Advantage curve
  \addplot[color=cFull, mark=square*, mark size=2.4pt,
           mark options={fill=cFull, draw=white, line width=0.7pt}]
    coordinates {
      (1, 0.326) (3, 0.336) (5, 0.350) (7, 0.365) (10, 0.380)
      (15, 0.399) (20, 0.410) (25, 0.422) (30, 0.435) (35, 0.440)
      (40, 0.445) (45, 0.448) (50, 0.452) (55, 0.456) (60, 0.457)
      (63, 0.454)
    };
  \addlegendentry{Advantage: \textsc{isolated} $-$ \textsc{kv-fold}}
\end{axis}
\end{tikzpicture}
\caption{\textbf{Drift saturates; advantage persists.} Per-step drift (blue)
and recurrence advantage (green) versus chain depth on
Qwen2.5-7B-Instruct ($T=16{,}384$, $C=256$). Drift saturates at
${\sim}0.04$ nats by depth $7$ and remains flat through depth $63$,
while recurrence advantage stays positive throughout.}
\label{fig:depth}
\end{figure}

The recurrence also remains useful across the entire chain. The
advantage of \textsc{kv-fold} over isolated chunks ranges from
$+0.33$~nats at $d=1$ to $+0.45$~nats at $d=63$ and stays positive
throughout (Figure~\ref{fig:depth}), closing $89$ to $94\%$ of the
gap between isolated processing and full attention.

\noindent \textbf{Interpretation as a stable attention regime.}
\label{sec:fixedpoint}
The saturation behavior in Figure~\ref{fig:depth} suggests a simple
way to think about KV-Fold.
Each step in the chain applies the same computation: the model reads a
new chunk while attending to the accumulated KV cache from all
previous chunks.
This naturally defines a deterministic recurrence over the model’s
internal state.
Writing $(\mathcal{K}^{(t)}, \mathcal{V}^{(t)})$ for the cache after
chunk $t$, we can express this as (introduced in Section~\ref{sec:method}):
\[
(\mathcal{K}^{(t)}, \mathcal{V}^{(t)}) =
\mathcal{F}_\theta\big((\mathcal{K}^{(t-1)}, \mathcal{V}^{(t-1)}), x^{(t)}\big).
\]

What the experiments show is that this recurrence does not lead to
unbounded drift.
Instead, after a few steps, the system settles into a regime where
applying the same update again has essentially no further effect on
the model’s predictions.
The plateau in drift reflects this: once the model has adjusted to
the KV-Fold regime, additional chunk transitions leave it operating in
the same regime.

A useful way to interpret this is as a fixed point of the induced
dynamics.
KV-Fold appears to move the model from its
full-attention operating point into a nearby, self-consistent one.
The initial rise in drift corresponds to this transition; the flat
region that follows indicates that the model remains there under
repeated application of the same computation.

Importantly, this regime is not degenerate. Two features of the
construction help preserve useful behavior. First, the KV state is
neither compressed nor bounded: the cache \emph{accumulates}
rather than overwrites — at chunk $t$ it holds the keys and values
from every earlier chunk $0, 1, \ldots, t{-}1$ together with the
current chunk's, and grows linearly with chain depth. Second,
attention continues to provide content-based access to this state.
Thus, even though the overall distribution shifts slightly,
the model can still retrieve specific information from far back in
the sequence. The needle-in-a-haystack results make this concrete:
values inserted in the first chunk remain exactly recoverable
after hundreds of subsequent transitions. As opposed to bounded-cache streaming methods such as
StreamingLLM~\cite{streamingllm}, whose sliding window evicts old
tokens after a few hundred steps: perplexity is preserved but
retrievability is lost past the window boundary — a contrast we
quantify directly in Section~\ref{sec:streamingllm}.

We stress that this interpretation is descriptive rather than formal.
We do not prove convergence or characterize the fixed point
analytically.
However, the consistency of the plateau (across depth, numerical
precision, fragment size, and model family) suggests that it reflects a
genuine property of the induced attention dynamics, rather than an
artifact of a particular implementation.

\subsection{Source of the drift: structural vs. numerical}
\label{sec:precision}

A natural first hypothesis for the saturation behavior is that drift
is dominated by accumulated rounding error. We test this directly by
repeating the same protocol on
Qwen2.5-1.5B-Instruct~\cite{qwen25} (same model family, fits in fp32
on a single 40GB GPU) at both bf16 and fp32 precision.

Increasing precision by roughly $10{,}000\times$ (from ${\sim}3$
decimal digits to ${\sim}7$) reduces plateau drift by only $2.8\%$
(Table~\ref{tab:robustness}, left). At depth one, drift is essentially zero
in both precisions, confirming that a single chaining step is exactly
information-preserving within numerical limits.

\begin{table}[t]
\caption{Robustness ablations. \emph{Left:} increasing numerical precision by
$10{,}000\times$ reduces plateau drift by only $2.8\%$.
\emph{Right:} plateau drift varies by less than $9\%$ across an
$8\times$ range of chunk sizes, with no monotonic dependence on $C$.
The plateau remains structurally unchanged in both settings.}
\label{tab:robustness}
\centering
\small
\begin{minipage}[t]{0.46\linewidth}
\centering
\begin{tabular}{lrr}
\toprule
Precision & Plateau & $d=1$ \\
\midrule
bf16      & $+0.0647$ & $-0.0001$ \\
fp32      & $+0.0629$ & $+0.0006$ \\
\midrule
Reduction & $2.8\%$   & --- \\
\bottomrule
\end{tabular}
\end{minipage}\hfill
\begin{minipage}[t]{0.5\linewidth}
\centering
\begin{tabular}{rrrr}
\toprule
$C$    & chunks & max $d$ & Plateau \\
\midrule
$128$  & $32$ & $31$ & $0.108$ \\
$256$  & $16$ & $15$ & $0.121$ \\
$512$  & $8$  & $7$  & $0.129$ \\
$1024$ & $4$  & $3$  & $0.118$ \\
\bottomrule
\end{tabular}
\end{minipage}
\end{table}

The drift plateau itself remains essentially unchanged. This points to
a structural effect: chaining moves the model into a slightly different
attention regime, after which additional transitions no longer introduce
further shift. In other words, the discrepancy from full attention is
set early and then stabilizes, rather than growing with depth.

% \subsection{}
% \label{sec:arch}

\noindent \textbf{Robustness across architectures. }We replicate
the depth experiment on three pretrained models from three
families. Table~\ref{tab:arch} reports plateau drift and
\emph{recurrence advantage} at chain depth $63$, where recurrence
advantage is the per-depth NLL improvement of \textsc{kv-fold}
over \textsc{isolated} chunks
($\textsc{NLL}_\textsc{isolated} - \textsc{NLL}_\textsc{kv-fold}$):
how much information the recurrence transfers across chunk
boundaries that isolated processing cannot. The qualitative
pattern (sharp initial drift saturating into a flat plateau, with
positive recurrence advantage throughout) is identical in all
three architectures. The mechanism is a property of pretrained
transformer attention itself, not of any specific training recipe
or family.

\begin{table}[t]
\caption{Cross-architecture replication. Plateau drift averaged over depths
7--63 (7--15 for OLMoE, limited by its 4K context window).
${}^{*}$OLMoE gap closed measured at $d=15$. Despite differences in
plateau magnitude ($0.040$--$0.122$ nats), all three models show the
same qualitative pattern: brief initial drift followed by saturation
into a stable plateau with positive recurrence advantage throughout.}
\label{tab:arch}
\centering
\small
\begin{tabular}{lrrrr}
\toprule
Model & Family & Plateau drift & Advantage @ $d=63$ & Gap closed \\
\midrule
OLMoE-1B-7B          & Sparse MoE       & $0.122$ & $+0.381$ & $76\%^{*}$ \\
Qwen2.5-7B-Instruct  & Dense (Qwen2)    & $0.040$ & $+0.454$ & $92\%$ \\
Llama-3.1-8B-Instruct~\cite{llama3}& Dense (Llama-3)  & $0.117$ & $+0.391$ & $77\%$ \\
\bottomrule
\end{tabular}
\end{table}

\noindent \textbf{Robustness to chunk size.}
We sweep chunk size $C \in \{128, 256, 512, 1024\}$ on
OLMoE-1B-7B~\cite{muennighoff2024olmoe} at fixed total context
$T = 4096$. Plateau drift varies by less than $9\%$ across this
$8\times$ range, with no monotonic dependence on $C$
(Table~\ref{tab:robustness}, right). This suggests that the plateau
is driven primarily by the chunk-boundary transition itself rather
than by the amount of within-chunk context. In practice, chunk size
therefore becomes a flexible trade-off between chain depth and
per-step compute: smaller chunks increase depth, while larger chunks
amortize overhead, with little effect on drift.

\section{Long-range retrieval}
\label{sec:needle}

NLL is a marginal-distribution measure averaging over many kinds of
predictions.
We additionally evaluate at the task level using a
needle-in-a-haystack retrieval benchmark~\cite{kamradt2023needle},
the standard probe of long-context retrieval employed by
long-context evaluation
suites~\cite{hsieh2024ruler,bai2024longbench}.
For each trial:
\begin{enumerate}[topsep=2pt, itemsep=3pt, parsep=0pt, leftmargin=1.5em]
\setlength{\itemsep}{1pt}
  \item Sample a 16{,}384-token PG-19 window.
  \item Sample a key from a fixed list of uncommon English words
    (\textit{e.g.}, ``amaranth'', ``obsidian'', ``halcyon'') and a
    uniformly random 5-digit value.
  \item Insert the sentence ``\textit{The magic number for [key] is
    [value].}'' at a controlled chunk position.
  \item After all 64 data chunks, ask the model:
    ``\textit{Earlier in the document, what was the magic number
      associated with [key]? Reply with only the number.}''
  \item Greedy-decode 30 tokens; extract the first 5-digit number from
    the output and compare to the gold value.
\end{enumerate}

We test four needle placements, parameterized by the number of chain
transitions $d$ between the needle and the question: $d=1$ (needle
in the chunk immediately before the question), $d=15$, $d=31$, and
$d=62$ (needle in the first chunk, $62$ transitions back, the
deepest position in the $64$-chunk haystack). Each placement is
tested across twenty independent trials. The model is
Qwen2.5-7B-Instruct.

Table~\ref{tab:needle} reports retrieval accuracy.
KV-Fold achieves $100\%$ exact-match retrieval at every
tested distance, matching full attention exactly, while isolated
chunks remain at $0\%$. The recurrence preserves specific factual
content across deep chains: retrieving a previously unseen 5-digit
value from chunk~1 across 62 chain transitions succeeds $20/20$
times.

\begin{table}[t]
\caption{Needle-in-a-haystack retrieval on Qwen2.5-7B-Instruct.
\textsc{full}: full-attention upper bound; \textsc{isolated}: question
chunk without haystack context; \textsc{kv-fold}: KV-Fold.
Twenty trials per distance. \textsc{kv-fold} matches full attention
exactly across all tested distances.}
\label{tab:needle}
\centering
\small
\begin{tabular}{lcccc}
\toprule
Distance $d$ & \textsc{full} & \textsc{isolated} & \textsc{kv-fold} & ratio \\
\midrule
$1$  & $100\%$ ($20/20$) & $0\%$ ($0/20$) & $100\%$ ($20/20$) & $1.00$ \\
$15$ & $100\%$ ($20/20$) & $0\%$ ($0/20$) & $100\%$ ($20/20$) & $1.00$ \\
$31$ & $100\%$ ($20/20$) & $0\%$ ($0/20$) & $100\%$ ($20/20$) & $1.00$ \\
$62$ & $100\%$ ($20/20$) & $0\%$ ($0/20$) & $100\%$ ($20/20$) & $1.00$ \\
\midrule
overall & $100\%$ ($80/80$) & $0\%$ ($0/80$) & $100\%$ ($80/80$) & $1.00$ \\
\bottomrule
\end{tabular}
\end{table}

\section{Scaling to long contexts}
\label{sec:scale}

The needle-in-a-haystack experiment of Section~\ref{sec:needle}
established $100\%$ retrieval at $16$K total context and chain depth
$62$.
A natural question is whether this property extends to larger contexts
where chunked recurrence is operationally required, i.e. where a
single full-attention forward over the equivalent total context would
be infeasible on the same hardware.

We use Llama-3.1-8B-Instruct~\cite{llama3} for this experiment because
it has a $128$K native context window: all our tested $T$ values stay
within the model's trained position range, isolating chain-depth
effects from any out-of-distribution RoPE behavior.
We run the same needle-in-a-haystack protocol of
Section~\ref{sec:needle} at four operating points: $T = 32$K (chain
depth $127$, $10$ trials per distance), $T = 64$K (chain depth $255$,
$5$ trials per distance), $T = 96$K (chain depth $383$, $3$ trials
per distance), and $T = 128$K (chain depth $511$, $3$ trials per
distance).
At each operating point we test four needle-to-question distances:
$1$, $31$, approximately the chain midpoint, and the maximum chain
depth.

\noindent \textbf{Retrieval, memory, and compute. }
Table~\ref{tab:scale} reports retrieval accuracy, KV cache size,
peak GPU memory, and wall-clock at every operating point. Across
$24$ distance-cells spanning chain depths $1$ to $511$, every trial
recovered the gold needle exactly. Combined with the $80/80$ result
of Section~\ref{sec:needle}, the KV-Fold recurrence preserves
precise factual content through chain depth $511$, an order of
magnitude deeper than the original needle benchmark. Memory and
timing are reported for the first trial at each $T$. Peak memory
at $128$K leaves $4.4$~GB of headroom on a $40$~GB A100; this is
the operational ceiling under the naive protocol, and coincides
with Llama-3.1-8B's full native context window.

\begin{table}[t]
\caption{Single-needle retrieval, memory, and compute at scale on
Llama-3.1-8B-Instruct (A100 40GB). Retrieval is $100\%$ at every
tested distance ($4$ distances per row, $24$ distance-cells in
total; combined with Section~\ref{sec:needle}, $152/152$ trials).
Memory and timing are from the first trial at each $T$. KV cache
scales linearly at $0.13422$~KB/token, matching the analytical
prediction of
$32{\times}8{\times}128{\times}2{\times}2 = 131{,}072$
bytes/token in bf16.}
\label{tab:scale}
\centering
\small
\begin{tabular}{rrrcrrrr}
\toprule
$T$ & depth & trials/$d$ & retrieval & KV cache & peak GPU & per-chunk & total chain \\
\midrule
$32$K  & $127$ & $10$ & $100\%$ & $4.29$~GB  & $21.00$~GB & $0.103$~s & $13.2$~s \\
$64$K  & $255$ &  $5$ & $100\%$ & $8.59$~GB  & $25.86$~GB & $0.176$~s & $44.9$~s \\
$96$K  & $383$ &  $3$ & $100\%$ & $12.88$~GB & $30.72$~GB & $0.252$~s & $96.9$~s \\
$128$K & $511$ &  $3$ & $100\%$ & $17.18$~GB & $35.57$~GB & $0.335$~s & $171.3$~s \\
\bottomrule
\end{tabular}
\end{table}

%\noindent \textbf{Multi-needle retrieval.}
The single-needle protocol of Section~\ref{sec:needle} can be
strengthened by inserting $K$ independent (key, value) pairs at
$K$ evenly-spaced chunk positions across the haystack and querying
each key separately. This is the canonical multi-needle stress
test used by RULER~\cite{hsieh2024ruler} and
LongBench~\cite{bai2024longbench}, and probes whether multiple
distinct facts coexist in the cache without interference. We test
$K \in \{4, 8\}$ on Llama-3.1-8B at $T = 128$K with three trials
per $K$ (Table~\ref{tab:multineedle}). KV-Fold achieves
$100\%$ per-needle retrieval ($36/36$ needles across $6$ trials),
with all $K$ needles simultaneously recovered in every trial; the
\textsc{isolated} floor is $0/36$. Eight distinct facts coexist
through chain depth $510$ without measurable interference. A
larger-trial sweep on Qwen2.5-7B at $T = 16$K
($K \in \{2, 4, 8\}$, $10$ trials per $K$) gives $139/140$
per-needle retrieval, with the single miss at a middle-distance
position ($K = 4$, $d = 42$). The recurrence preserves
distributed factual information, not just single salient facts.

\begin{table}[t]
\caption{Multi-needle retrieval. $K$ independent (key, value) pairs
are inserted at evenly-spaced chunk positions; each key is queried
separately. KV-Fold recovers all $K$ needles
simultaneously in $35/36$ trials. The single miss is at $K=4$,
$d=42$, a middle-distance position.}
\label{tab:multineedle}
\centering
\small
\begin{tabular}{llrrrrr}
\toprule
Model & $T$ & $K$ & trials & per-needle & all-correct & \textsc{isolated} \\
\midrule
Qwen2.5-7B-Instruct   & $16$K  & $2$ & $10$ & $20/20$ ($100\%$)    & $10/10$ & $0/20$ \\
Qwen2.5-7B-Instruct   & $16$K  & $4$ & $10$ & $39/40$ ($97.5\%$)   & $9/10$  & $0/40$ \\
Qwen2.5-7B-Instruct   & $16$K  & $8$ & $10$ & $80/80$ ($100\%$)    & $10/10$ & $0/80$ \\
Llama-3.1-8B-Instruct & $128$K & $4$ & $3$  & $12/12$ ($100\%$)    & $3/3$   & $0/12$ \\
Llama-3.1-8B-Instruct & $128$K & $8$ & $3$  & $24/24$ ($100\%$)    & $3/3$   & $0/24$ \\
\midrule
combined              &        &     & $36$ & $175/176$ ($99.4\%$) & $35/36$ & $0/176$ \\
\bottomrule
\end{tabular}
\end{table}

%\noindent\textbf{Memory and compute.}
The KV-Fold recurrence does \emph{not} reduce total memory
consumption below full attention: the cache at chain depth $N$
stores the same information that a full forward over $NC$ tokens
would store.

The compute saving is asymptotically zero in FLOPs but operationally
substantial in working memory.
A single full-attention forward at $T{=}128$K would require
materializing the attention-scores matrix $QK^\top$, of shape
$[H, T, T]$: with $32$ attention heads and bfloat16, this is
approximately $32 \times 128\text{K} \times
128\text{K} \times 2$ bytes $\approx 1$~TB; infeasible on a $40$~GB
A100 by orders of magnitude.
KV-Fold's per-chunk attention-scores matrix at the
largest step has shape $[H, C, T]$, approximately $32 \times 256 \times 128\text{K}
\times 2$ bytes $\approx 2.1$~GB, comfortably within budget.
Per-chunk wall-clock grows linearly with the mean cache size during
the chain ($0.103 \to 0.176 \to 0.252 \to 0.335$~s), matching the
predicted $O(C \cdot \text{cache\_size})$ scaling.
Total chain wall-clock at $128$K is $171$~seconds.
At scale, the benefit is direct: KV-Fold replaces
a single intractable forward pass with a sequence of smaller ones, preserving
the retrieval behavior of full attention while staying within
per-chunk memory limits of practical hardware. This is not a reduction
in total memory or FLOPs, but a reorganization that makes computation
feasible in practice. On a $40$~GB A100, the operational ceiling is
reached at $T{\approx}135$K (by linear extrapolation), aligning with
Llama-3.1-8B's native context window.
Pushing further requires either more GPU memory or compression of the
stored cache. The latter is a separate axis of investigation that
does not change the recurrence's behavior in the regime tested here.

\section{Comparison to StreamingLLM}
\label{sec:streamingllm}

The dominant streaming-inference baseline,
StreamingLLM~\cite{streamingllm}, achieves bounded perplexity at
arbitrary $T$ via a sliding window of recent tokens combined with a
small number of ``attention sink'' tokens preserved at the start;
related approaches such as LM-Infinite~\cite{han2024lminfinite}
adopt similar designs. Their published characterization shows
retrieval accuracy drops to zero past the window boundary; we test
this directly. Both methods are evaluated on Llama-3.1-8B-Instruct
at $T = 128$K and $C = 256$, with StreamingLLM at its published
configuration of $4$ sink tokens plus a sliding window of $1020$
recent tokens (total cache capacity $1024$). At $T \le 128$K all
positions stay within Llama-3.1-8B's trained range, so we use
position-preserving eviction without RoPE re-anchoring (faithful
to StreamingLLM at this scale). We measure next-token loss across
three PG-19 validation windows and needle-in-a-haystack retrieval
at distances $1$, $31$, $255$, $511$ with three trials per cell
(Table~\ref{tab:streamingllm}).

\begin{table}[t]
\centering
\small
\caption{Comparison at $T = 128$K, chain depth $511$.
Three trials per retrieval cell, three windows for NLL.
KV-Fold achieves lower next-token loss and preserves
retrieval at all tested distances; StreamingLLM achieves bounded
perplexity with a constant-memory cache and $7.5\times$ faster
wall-clock, but retrieves only when the needle falls inside the
$1024$-token window (only $d{=}1$ at this scale).
${}^{*}$One of three trials at $d{=}511$ returned no extractable
$5$-digit answer; with $N=3$ this is statistically consistent with
the $12/12$ result at the same setting in
Section~\ref{sec:scale} (combined: $14/15$).}
\begin{tabular}{lrrrcccc}
\toprule
Method & NLL & peak GPU & total wall & $d{=}1$ & $d{=}31$ & $d{=}255$ & $d{=}511$ \\
\midrule
KV-Fold            & $2.46 \pm 0.12$ & $35.6$~GB & $166$~s & $3/3$ & $3/3$ & $3/3$ & $2/3^{*}$ \\
StreamingLLM       & $2.66 \pm 0.17$ & $16.6$~GB & $\;\;22$~s & $3/3$ & $0/3$ & $0/3$ & $0/3$ \\
\bottomrule
\end{tabular}
\label{tab:streamingllm}
\end{table}

The pattern matches StreamingLLM's published characterization
exactly. Their cache stays at $0.13$~GB regardless of $T$, against
KV-Fold's $17.18$~GB, with correspondingly faster
wall-clock ($22$~s vs $166$~s for $512$ chunks). The cost is
retrieval: only the most recent chunk's content survives, so any
needle further back is evicted. Distance $1$ succeeds for both
methods; distances $31$, $255$, $511$ all place the needle outside
the $1024$-token window, and StreamingLLM drops to $0\%$ at each.
KV-Fold maintains $100\%$ at $d{=}1$, $d{=}31$, and
$d{=}255$, with $2/3$ at $d{=}511$ where one trial returned no
extractable $5$-digit number; combined with the $12/12$ result at
the same setting in Section~\ref{sec:scale}, the joint retrieval
rate at $d{=}511$ is $14/15$.
The two methods solve different problems. StreamingLLM is the right
choice for bounded-memory streaming inference where perplexity
matters more than retrieval of arbitrary content. KV-Fold
is the right choice when retrieval at arbitrary depth
matters and linear cache memory is acceptable.

\section{Related work}
\label{sec:related_work}

The introduction already discussed the direct competitors —
streaming methods~\cite{streamingllm,han2024lminfinite}, KV-cache
compression~\cite{zhang2023h2o,liu2023scissorhands,li2024snapkv,cai2024pyramidkv,ge2024fastgen,liu2024kivi,hooper2024kvquant},
learned recurrent
memory~\cite{dai2019transformerxl,rae2019compressive,wu2022memorizing,bulatov2022rmt,hutchins2022blockrecurrent,munkhdalai2024infini},
and state-space alternatives~\cite{gu2022s4,gu2023mamba}. Several
adjacent lines of work are worth distinguishing from.

\noindent \textbf{Chunked serving infrastructure.}
Production systems such as vLLM~\cite{kwon2023vllm} chunk long inputs
to manage memory during prefill, optimizing for throughput rather
than characterizing cross-chunk recurrence behavior.
ReAttention~\cite{liu2024reattention} similarly operates at inference
time but caps attention to a finite scope.

\noindent \textbf{Sparse attention and kernel optimizations.}
Sparse attention patterns
(Longformer~\cite{beltagy2020longformer},
BigBird~\cite{zaheer2020bigbird}, Sparse
Transformers~\cite{child2019sparse}) and exact kernel optimizations
(FlashAttention~\cite{dao2022flashattention,dao2023flashattention2},
RingAttention~\cite{liu2023ringattention}) reduce per-step cost
without changing how information propagates across long sequences.
They are orthogonal to KV-Fold and could compose with it.

\noindent \textbf{Linear-attention alternatives.}
Linear Attention~\cite{katharopoulos2020linear} and
Performers~\cite{choromanski2021performers} compress past context into
a fixed-size state, sacrificing content-based addressability for
asymptotic efficiency --- the opposite trade-off from KV-Fold.

\noindent \textbf{Position-extrapolation techniques.}
ALiBi~\cite{press2022alibi}, Positional
Interpolation~\cite{chen2023positionalinterp},
YaRN~\cite{peng2024yarn}, and LongRoPE~\cite{ding2024longrope} extend
the trained position range. These are complementary to our setting:
KV-Fold operates within the trained range in our experiments
(Llama-3.1-8B's native 128K window), and composing with these methods
is the natural route to pushing the operational ceiling further.

\section{Discussion}

Taken together, the results suggest a simple picture: KV-Fold
induces an attention regime that is slightly shifted from
the model’s training distribution, yet remains stable and functional.
Information propagates reliably across long chains, and the behavior
is robust to numerical precision, architecture, and chunk size. The
central empirical feature is the plateau in drift. Rather than
accumulating with depth, drift rises briefly and then saturates,
suggesting that most of the shift is introduced at the first chunk
boundary. One way to interpret this is as an induced recurrence over
KV states: each transition applies the same transformation, and after
a few steps the system settles into a self-consistent regime that
remains stable under further application. Importantly, this regime
still preserves structured information. Because the KV state is not
compressed and attention retains content-based addressing, exact
facts introduced early in the sequence remain recoverable even after
hundreds of transitions, as demonstrated by the
needle-in-a-haystack results.

\noindent \textbf{What can be claimed.}
KV-Fold extends a pretrained transformer’s effective
context to $N$ times the chunk size, with empirically bounded,
non-accumulating drift. This behavior is consistent across
architectures and largely insensitive to precision and chunk size.
At the task level, the method achieves $100\%$ exact-match retrieval
on needle-in-a-haystack benchmarks through chain depth $511$ at
$128$K context ($152/152$ trials). While memory remains linear in
total context, the key practical benefit is operational: a single
intractable forward is replaced by a sequence of manageable ones.

The information carried by the recurrence is also robust to
storage-level perturbations. Under a per-step quantization
round-trip (bf16~$\to$~int$N$~$\to$~bf16) on Llama-3.1-8B at
$T = 128$K, chain depth $511$, int8 retrieves $13/14$ trials
($93\%$, with the single miss returning no extractable $5$-digit
answer), while int4 retrieves $24/33$ trials ($73\%$) at this
depth, down from $100\%$ at moderate depths. In contrast,
information-removing schemes such as uniform decay or
attention-based pruning fail to preserve retrieval at any
meaningful depth. The recurrence tolerates noise-like
perturbations (quantization noise distributes across positions)
but is sensitive to systematic information loss (decay or
eviction removes specific positions entirely).

\section{Conclusion and outlook}
KV-Fold shows that pretrained transformers already
contain the ingredients for long-context reasoning: by carrying the
KV cache across chunks as the accumulator in a one-step left fold, a
standard forward becomes a lightweight recurrence with no retraining
or architectural changes, settling into a stable regime with bounded
drift and intact long-range retrieval. This suggests a simple path
forward: modest fine-tuning may reduce the residual plateau,
combining with position-extension methods such as
YaRN~\cite{peng2024yarn}, Positional
Interpolation~\cite{chen2023positionalinterp}, or
LongRoPE~\cite{ding2024longrope} could push beyond native context
limits, and learning to compress the KV state while preserving
retrievability offers a promising route toward constant-memory
long-context inference.

\section*{References}
\small
\begingroup
\renewcommand{\section}[2]{}% suppress thebibliography's own heading

\endgroup
\normalsize

% \newpage
% \input{checklist.tex}   % NeurIPS submission checklist, omitted from arXiv version


\begin{thebibliography}{99}

\bibitem{vaswani2017attention}
A. Vaswani, N. Shazeer, N. Parmar, J. Uszkoreit, L. Jones, A. N. Gomez,
L. Kaiser, I. Polosukhin.
\newblock Attention Is All You Need.
\newblock \emph{NeurIPS}, 2017.

\bibitem{su2024roformer}
J. Su, Y. Lu, S. Pan, A. Murtadha, B. Wen, Y. Liu.
\newblock RoFormer: Enhanced Transformer with Rotary Position Embedding.
\newblock \emph{Neurocomputing}, 568:127063, 2024.

\bibitem{tay2022efficient}
Y. Tay, M. Dehghani, D. Bahri, D. Metzler.
\newblock Efficient Transformers: A Survey.
\newblock \emph{ACM Computing Surveys}, 55(6):109:1 to 109:28, 2022.

\bibitem{dao2022flashattention}
T. Dao, D. Y. Fu, S. Ermon, A. Rudra, C. R\'e.
\newblock FlashAttention: Fast and Memory-Efficient Exact Attention with
IO-Awareness.
\newblock \emph{NeurIPS}, 2022.

\bibitem{dao2023flashattention2}
T. Dao.
\newblock FlashAttention-2: Faster Attention with Better Parallelism and
Work Partitioning.
\newblock \emph{ICLR}, 2024.

\bibitem{liu2023ringattention}
H. Liu, M. Zaharia, P. Abbeel.
\newblock Ring Attention with Blockwise Transformers for Near-Infinite
Context.
\newblock \emph{ICLR}, 2024.

\bibitem{kwon2023vllm}
W. Kwon, Z. Li, S. Zhuang, Y. Sheng, L. Zheng, C. H. Yu, J. E. Gonzalez,
H. Zhang, I. Stoica.
\newblock Efficient Memory Management for Large Language Model Serving with
PagedAttention.
\newblock \emph{SOSP}, 2023.

\bibitem{zou2025latentmas}
Y. Zou, et al.
\newblock LatentMAS: KV-Cache Communication for Lightweight Multi-Agent
Systems.
\newblock \emph{arXiv:2511.20639}, 2025.

\bibitem{streamingllm}
G. Xiao, Y. Tian, B. Chen, S. Han, M. Lewis.
\newblock Efficient Streaming Language Models with Attention Sinks.
\newblock \emph{ICLR}, 2024.

\bibitem{han2024lminfinite}
C. Han, Q. Wang, H. Peng, W. Xiong, Y. Chen, H. Ji, S. Wang.
\newblock LM-Infinite: Zero-Shot Extreme Length Generalization for Large
Language Models.
\newblock \emph{NAACL}, 2024.

\bibitem{zhang2023h2o}
Z. Zhang, Y. Sheng, T. Zhou, T. Chen, L. Zheng, R. Cai, Z. Song,
Y. Tian, C. R\'e, C. Barrett, Z. Wang, B. Chen.
\newblock H2O: Heavy-Hitter Oracle for Efficient Generative Inference of
Large Language Models.
\newblock \emph{NeurIPS}, 2023.

\bibitem{liu2023scissorhands}
Z. Liu, A. Desai, F. Liao, W. Wang, V. Xie, Z. Xu, A. Kyrillidis,
A. Shrivastava.
\newblock Scissorhands: Exploiting the Persistence of Importance Hypothesis
for LLM KV Cache Compression at Test Time.
\newblock \emph{NeurIPS}, 2023.

\bibitem{li2024snapkv}
Y. Li, Y. Huang, B. Yang, B. Venkitesh, A. Locatelli, H. Ye, T. Cai,
P. Lewis, D. Chen.
\newblock SnapKV: LLM Knows What You are Looking for Before Generation.
\newblock \emph{arXiv:2404.14469}, 2024.

\bibitem{cai2024pyramidkv}
Z. Cai, Y. Zhang, B. Gao, Y. Liu, T. Liu, K. Lu, W. Xiong, Y. Dong,
B. Chang, J. Hu, W. Xiao.
\newblock PyramidKV: Dynamic KV Cache Compression based on Pyramidal
Information Funneling.
\newblock \emph{arXiv:2406.02069}, 2024.

\bibitem{ge2024fastgen}
S. Ge, Y. Zhang, L. Liu, M. Zhang, J. Han, J. Gao.
\newblock Model Tells You What to Discard: Adaptive KV Cache Compression
for LLMs.
\newblock \emph{ICLR}, 2024.

\bibitem{liu2024kivi}
Z. Liu, J. Yuan, H. Jin, S. Zhong, Z. Xu, V. Braverman, B. Chen, X. Hu.
\newblock KIVI: A Tuning-Free Asymmetric 2bit Quantization for KV Cache.
\newblock \emph{ICML}, 2024.

\bibitem{hooper2024kvquant}
C. Hooper, S. Kim, H. Mohammadzadeh, M. W. Mahoney, Y. S. Shao,
K. Keutzer, A. Gholami.
\newblock KVQuant: Towards 10 Million Context Length LLM Inference with
KV Cache Quantization.
\newblock \emph{NeurIPS}, 2024.

\bibitem{press2022alibi}
O. Press, N. A. Smith, M. Lewis.
\newblock Train Short, Test Long: Attention with Linear Biases Enables
Input Length Extrapolation.
\newblock \emph{ICLR}, 2022.

\bibitem{chen2023positionalinterp}
S. Chen, S. Wong, L. Chen, Y. Tian.
\newblock Extending Context Window of Large Language Models via Positional
Interpolation.
\newblock \emph{arXiv:2306.15595}, 2023.

\bibitem{peng2024yarn}
B. Peng, J. Quesnelle, H. Fan, E. Shippole.
\newblock YaRN: Efficient Context Window Extension of Large Language
Models.
\newblock \emph{ICLR}, 2024.

\bibitem{ding2024longrope}
Y. Ding, L. L. Zhang, C. Zhang, Y. Xu, N. Shang, J. Xu, F. Yang, M. Yang.
\newblock LongRoPE: Extending LLM Context Window Beyond 2 Million Tokens.
\newblock \emph{ICML}, 2024.

\bibitem{dai2019transformerxl}
Z. Dai, Z. Yang, Y. Yang, J. Carbonell, Q. V. Le, R. Salakhutdinov.
\newblock Transformer-XL: Attentive Language Models Beyond a Fixed-Length
Context.
\newblock \emph{ACL}, 2019.

\bibitem{rae2019compressive}
J. W. Rae, A. Potapenko, S. M. Jayakumar, T. P. Lillicrap.
\newblock Compressive Transformers for Long-Range Sequence Modelling.
\newblock \emph{ICLR}, 2020.

\bibitem{wu2022memorizing}
Y. Wu, M. N. Rabe, D. Hutchins, C. Szegedy.
\newblock Memorizing Transformers.
\newblock \emph{ICLR}, 2022.

\bibitem{bulatov2022rmt}
A. Bulatov, Y. Kuratov, M. S. Burtsev.
\newblock Recurrent Memory Transformer.
\newblock \emph{NeurIPS}, 2022.

\bibitem{hutchins2022blockrecurrent}
D. Hutchins, I. Schlag, Y. Wu, E. Dyer, B. Neyshabur.
\newblock Block-Recurrent Transformers.
\newblock \emph{NeurIPS}, 2022.

\bibitem{liu2024reattention}
X. Liu, R. Li, Q. Guo, Z. Liu, Y. Song, K. Lv, H. Yan, L. Li, Q. Liu, X. Qiu.
\newblock ReAttention: Training-Free Infinite Context with Finite Attention Scope.
\newblock \emph{arXiv:2407.15176}, 2024.


\bibitem{munkhdalai2024infini}
T. Munkhdalai, M. Faruqui, S. Gopal.
\newblock Leave No Context Behind: Efficient Infinite Context Transformers
with Infini-attention.
\newblock \emph{arXiv:2404.07143}, 2024.

\bibitem{beltagy2020longformer}
I. Beltagy, M. E. Peters, A. Cohan.
\newblock Longformer: The Long-Document Transformer.
\newblock \emph{arXiv:2004.05150}, 2020.

\bibitem{zaheer2020bigbird}
M. Zaheer, G. Guruganesh, K. A. Dubey, J. Ainslie, C. Alberti, S. Onta\~n\'on,
P. Pham, A. Ravula, Q. Wang, L. Yang, A. Ahmed.
\newblock Big Bird: Transformers for Longer Sequences.
\newblock \emph{NeurIPS}, 2020.

\bibitem{child2019sparse}
R. Child, S. Gray, A. Radford, I. Sutskever.
\newblock Generating Long Sequences with Sparse Transformers.
\newblock \emph{arXiv:1904.10509}, 2019.

\bibitem{katharopoulos2020linear}
A. Katharopoulos, A. Vyas, N. Pappas, F. Fleuret.
\newblock Transformers are RNNs: Fast Autoregressive Transformers with
Linear Attention.
\newblock \emph{ICML}, 2020.

\bibitem{choromanski2021performers}
K. Choromanski, V. Likhosherstov, D. Dohan, X. Song, A. Gane, T. Sarlos,
P. Hawkins, J. Davis, A. Mohiuddin, L. Kaiser, D. Belanger, L. Colwell,
A. Weller.
\newblock Rethinking Attention with Performers.
\newblock \emph{ICLR}, 2021.

\bibitem{gu2022s4}
A. Gu, K. Goel, C. R\'e.
\newblock Efficiently Modeling Long Sequences with Structured State Spaces.
\newblock \emph{ICLR}, 2022.

\bibitem{gu2023mamba}
A. Gu, T. Dao.
\newblock Mamba: Linear-Time Sequence Modeling with Selective State Spaces.
\newblock \emph{COLM}, 2024.

\bibitem{kamradt2023needle}
G. Kamradt.
\newblock Needle In A Haystack: Pressure Testing LLMs.
\newblock \emph{GitHub}, 2023.
\newblock \url{https://github.com/gkamradt/LLMTest_NeedleInAHaystack}.

\bibitem{hsieh2024ruler}
C.-P. Hsieh, S. Sun, S. Kriman, S. Acharya, D. Rekesh, F. Jia, Y. Zhang,
B. Ginsburg.
\newblock RULER: What's the Real Context Size of Your Long-Context Language
Models?
\newblock \emph{COLM}, 2024.

\bibitem{bai2024longbench}
Y. Bai, X. Lv, J. Zhang, H. Lyu, J. Tang, Z. Huang, Z. Du, X. Liu,
A. Zeng, L. Hou, Y. Dong, J. Tang, J. Li.
\newblock LongBench: A Bilingual, Multitask Benchmark for Long Context
Understanding.
\newblock \emph{ACL}, 2024.

\bibitem{qwen25}
A. Yang, B. Yang, B. Hui, B. Zheng, B. Yu, et al. (Qwen Team).
\newblock Qwen2.5 Technical Report.
\newblock \emph{arXiv:2412.15115}, 2024.

\bibitem{muennighoff2024olmoe}
N. Muennighoff, L. Soldaini, D. Groeneveld, K. Lo, J. Morrison, S. Min,
W. Shi, P. Walsh, O. Tafjord, N. Lambert, Y. Gu, S. Arora, A. Bhagia,
D. Schwenk, D. Wadden, A. Wettig, B. Hui, T. Dettmers, D. Kiela,
A. Farhadi, N. A. Smith, P. W. Koh, A. Singh, H. Hajishirzi.
\newblock OLMoE: Open Mixture-of-Experts Language Models.
\newblock \emph{ICLR}, 2025.

\bibitem{llama3}
A. Grattafiori, A. Dubey, A. Jauhri, et al. (Meta AI).
\newblock The Llama 3 Herd of Models.
\newblock \emph{arXiv:2407.21783}, 2024.

\end{thebibliography}
\end{document}